\documentclass[twocolumn,letterpaper, 12pt]{article}
\usepackage{import}
\usepackage{epa}
\usepackage[left=.65in,top=.65in,bottom=.65in,right=.65in]{geometry}
\usepackage[hidelinks]{hyperref}
\usepackage{booktabs}
\usepackage{dblfloatfix}

\title{The Effects of Synthetic Data and Label Distribution on Canola Branch Counting}

\author{%
\textbf{Amirsalar Darvishpour, Mikolaj Cieslak, and Adam Runions}\\
\begin{small}
Department of Computer Science, University of Calgary \\
\end{small}
}

\date{}
\begin{document}
\maketitle

%%%%%%%%%%%%%%%%%%%%%%%%%%%%%%%%%%%%%%%%%%%%%%%%%%%%%%%%%%%%%%%%%%%%%
\section*{Abstract}
%%%%%%%%%%%%%%%%%%%%%%%%%%%%%%%%%%%%%%%%%%%%%%%%%%%%%%%%%%%%%%%%%%%%%

Collecting annotated plant images for automated phenotyping is often slow and expensive. Plant models simulating growth and development can generate unlimited
synthetic images with exact labels. However, previous work has established that
whether incorporating synthetic data improves performance depends on the ratio
of synthetic to real images and the label distribution of the synthetic
dataset~\cite{ubbens2018use}.
To systematically quantify both factors, we train ResNet-18 models on a canola
branch-counting task using a calibrated L-system plant
model~\cite{cieslak2021lsystem}.
We vary each factor independently.
Synthetic-to-real ratios of 1:5 to 1:22 broadly improve performance; the best
ratio (1:7) reduces mean absolute difference by 7.6\% over real-only training.
For label distribution, a uniform synthetic distribution is strongly suboptimal
(abs.\ diff.\ $\approx$1.70); interpolating 90\% toward the real distribution
yields abs.\ diff.\ 0.927, whereas Gaussian smoothing of the real label
distribution yields the best overall result (abs.\ diff.\ 0.912, a 14.7\%
improvement over real-only).
A minimum of 10 synthetic images per label offers a simpler alternative with
modest gains, while 100 per label over-corrects and hurts performance.

\begin{EPAKeywords}
L-system, synthetic data augmentation, plant phenotyping, label distribution, ResNet
\end{EPAKeywords}

%%%%%%%%%%%%%%%%%%%%%%%%%%%%%%%%%%%%%%%%%%%%%%%%%%%%%%%%%%%%%%%%%%%%%
\section{Introduction}
%%%%%%%%%%%%%%%%%%%%%%%%%%%%%%%%%%%%%%%%%%%%%%%%%%%%%%%%%%%%%%%%%%%%%

Automated plant phenotyping via supervised deep learning often requires large annotated
datasets, which are costly to produce~\cite{walter2015plant}.
Procedural L-system plant models can generate unlimited synthetic images with
exact organ-count labels~\cite{ubbens2018use,cieslak2021lsystem}, and have been 
used in ML model training for plant counting tasks in
Arabidopsis~\cite{ubbens2018use}, canola~\cite{khan2025effectiveness}, and
wheat~\cite{beheshtifard2025beyond}.
However, due to the domain gap between synthetic and real images, how synthetic
data is incorporated has a substantial impact on model performance.

Previous work has identified two particularly important factors for dataset
design~\cite{ubbens2018use,khan2025effectiveness}: (i) the ratio of synthetic to
real images and (ii) the label distribution of the synthetic dataset.
Our work builds on initial explorations by Khan et al.~\cite{khan2025effectiveness},
who used the same canola dataset to investigate how the amount of real images
affects performance.
By contrast, we focus on synthetic data design and perform a quantitative
characterization of the impact of synthetic-to-real ratio and label distribution
over a wider range of conditions than previous studies.
We identify a broad range of ratios where the addition of synthetic data improves
performance.
With respect to distributions, we examine the impact of shifting the synthetic
label distribution from uniform toward the real data distribution, as well as
the effect of Gaussian smoothing applied to the real distribution.
Notably, we find that shifting the synthetic distribution away from the actual,
observed distribution by smoothing or slightly increasing the representation of 
under-represented labels improves model performance.

%%%%%%%%%%%%%%%%%%%%%%%%%%%%%%%%%%%%%%%%%%%%%%%%%%%%%%%%%%%%%%%%%%%%%
\section{Materials and Methods}
%%%%%%%%%%%%%%%%%%%%%%%%%%%%%%%%%%%%%%%%%%%%%%%%%%%%%%%%%%%%%%%%%%%%%

\begin{figure}[!h]
  \centering
  \includegraphics[width=\columnwidth]{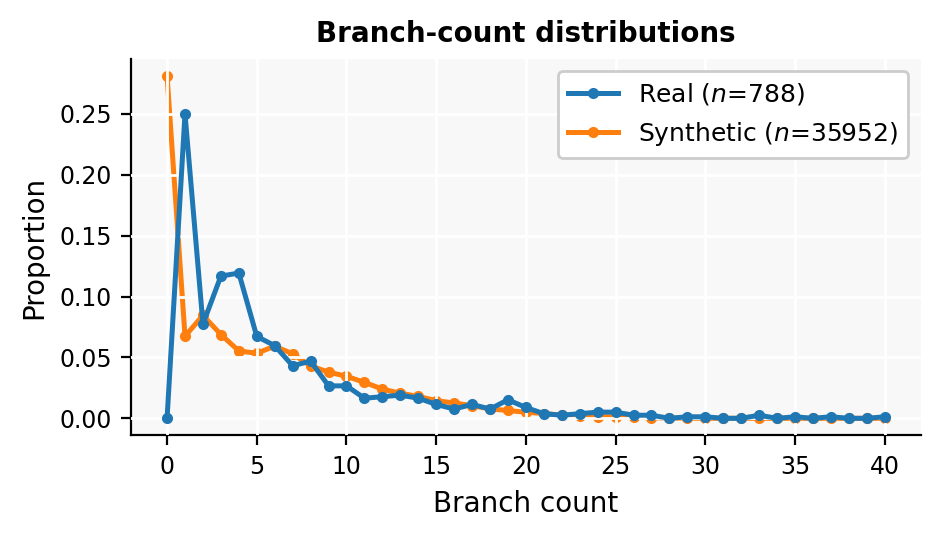}
  \caption{Branch-count distributions of the real (blue) and synthetic
           (orange) datasets.
           The synthetic distribution arises from the L-system model
           calibrated by Khan et al.~\cite{khan2025effectiveness}.}
  \label{fig:distributions}
\end{figure}

\textbf{Dataset.}
A dataset of 788 real canola images with ground-truth branch counts (range 1--40)
was collected in a high-throughput indoor phenotyping facility~\cite{ebersbach2021exploiting}.
Synthetic images were generated by a calibrated L-system canola
model~\cite{cieslak2021lsystem}; Khan et al.~\cite{khan2025effectiveness}
calibrated the model parameters to qualitatively match the synthetic branch counts to the real ones.
Each plant is rendered at a randomly selected growth day (day 27--50), and the
resulting variation in branch count across growth stages defines the
\textit{synthetic distribution} used in the ratio sweep experiment.
Figure~\ref{fig:distributions} shows the branch-count distributions of both datasets.

\textbf{Model.}
A ResNet-18~\cite{he2016deep} pre-trained on ImageNet was fine-tuned for branch
count regression (Adam, lr\,$=\!10^{-4}$, batch 16, early stopping patience 30).
The 788 real images were split into 400 training, 100 validation, and 288 test
images; all experiments were repeated five times and evaluated on the held-out
test set, with results reported as mean\,$\pm$\,std of the mean absolute
difference (abs.\ diff.).

\textbf{Exp.\ 1 --- Ratio sweep.}
A fixed real set was combined with synthetic images at thirty ratios
($1\!:\!N$, $N\!\in\![0,29]$, real:synthetic).

\textbf{Exp.\ 2 --- Distribution interpolation.}
The synthetic label distribution was shifted from uniform (Step~0) to the real
label distribution (Step~10) in ten equal steps (each moving 10\% closer),
using 18k or 36k synthetic images paired with the 400 real training images;
the target distribution is computed from all 788 real images (45:1 ratio at
18k, outside the optimal range from Exp.~1, to isolate the effect of
distribution).
Images were selected from a large pre-generated pool targeting a specified
set of branch-count frequencies.
The expectation that slightly over-representing rare labels improves performance
motivates Experiments~3 and~4.

\textbf{Exp.\ 3 --- Gaussian smoothing.}
A Gaussian kernel was convolved with the real label distribution histogram to
smooth branch-count frequencies across neighbouring values; ten values of
$\sigma$ ($0.0, 0.4, \ldots, 3.6$) evenly divide the interval $[0, 4)$ in
steps of $0.4$.

\textbf{Exp.\ 4 --- Minimum per label.}
Hard-floor constraints of 10 (min-10) and 100 (min-100) synthetic images per
label value were tested, up-sampling rare high-count labels.
This experiment tests a simpler alternative to distribution resampling that
ensures basic label coverage without requiring knowledge of the target distribution.

\begin{figure}[!tbph]
  \centering
  \includegraphics[width=\columnwidth]{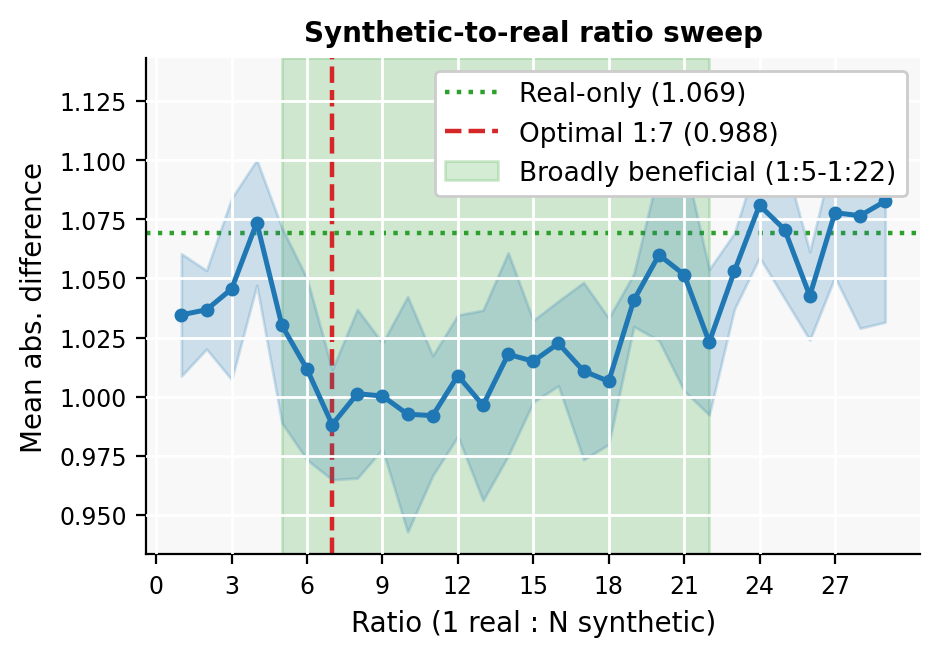}
  \caption{Abs.\ diff.\ vs.\ synthetic-to-real ratio ($N$ in 1:$N$).
           Green shading: broadly beneficial range (1:5--1:22).
           Dotted green: real-only baseline (1.069).
           Dashed red: optimal ratio 1:7 (0.988).
           Error bars: $\pm$1\,std, five runs.}
  \label{fig:ratio}
\end{figure}

\begin{figure*}[!tb]
  \centering
  \includegraphics[width=\textwidth]{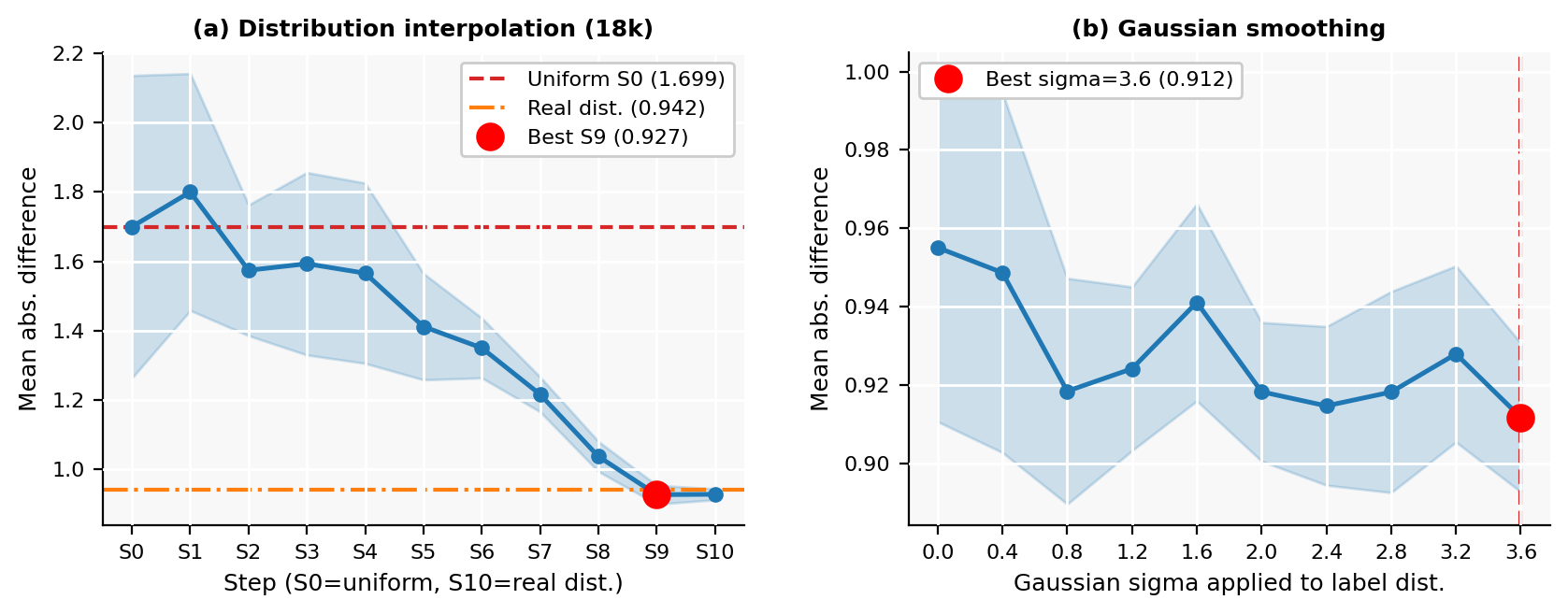}
  \caption{\textbf{(a)} Abs.\ diff.\ vs.\ distribution interpolation step
           (S0\,=\,uniform; S10\,=\,real distribution), 18k synthetic images.
           Dashed red: uniform baseline; orange dash-dot: real distribution;
           red circle: best step (S9, 0.927).
           \textbf{(b)} Abs.\ diff.\ vs.\ Gaussian smoothing $\sigma$.
           Red circle: best $\sigma$ (3.6, abs.\ diff.\,=\,0.912).
           Error bars: $\pm$1\,std, five runs.}
  \label{fig:distribution}
\end{figure*}

%%%%%%%%%%%%%%%%%%%%%%%%%%%%%%%%%%%%%%%%%%%%%%%%%%%%%%%%%%%%%%%%%%%%%
\section{Results and Discussion}
%%%%%%%%%%%%%%%%%%%%%%%%%%%%%%%%%%%%%%%%%%%%%%%%%%%%%%%%%%%%%%%%%%%%%

\subsection{Experiment 1: Ratio Sweep}

Figure~\ref{fig:ratio} shows abs.\ diff.\ across the thirty ratios.
Real-only training (1:0) achieves 1.069.
Most ratios from 1:1 to approximately 1:22 outperform this baseline, with
significant improvement visible from 1:5 onward.
The best single ratio is 1:7 (abs.\ diff.\,=\,0.988, $-7.6\%$).
Performance is relatively flat in the 1:7--1:18 window (0.988--1.017), so
practitioners need not tune the ratio precisely within this range.
Beyond 1:22, abs.\ diff.\ exceeds the real-only baseline, showing that
excess synthetic data amplifies domain-gap noise.
This extends the findings of Ubbens et al.~\cite{ubbens2018use} to canola,
identifying a broadly beneficial ratio range as well as the point at which
synthetic data becomes detrimental.

\subsection{Experiments 2 \& 3: Label Distribution}

Figure~\ref{fig:distribution} shows results for distribution interpolation
and Gaussian smoothing.

\textbf{Interpolation (panel~a).}
A uniform distribution (Step~0) is dramatically suboptimal
(abs.\ diff.\,=\,1.700), consistent with Ubbens et al.~\cite{ubbens2018use},
who observed that distributional mismatch degrades generalisation.
Error decreases monotonically from Step~0 to Step~9, with the sharpest gains
between Steps 5 and 9.
The best result is at Step~9 (0.927), outperforming even the real distribution
(0.942), suggesting two benefits of moving slightly toward uniform: it smooths
out frequency extremes and increases the minimum count of rare labels.
Gaussian smoothing (Exp.~3) captures the first effect and minimum-per-label
floors (Exp.~4) capture the second.

With 36k synthetic images, Step~9 was still the best condition (1.211), but all
results were substantially worse than their 18k counterparts.
This confirms that even a well-matched distribution does not compensate for an
excess of synthetic data---the ratio remains a critical factor.

\textbf{Gaussian smoothing (panel~b).}
Starting from the real label distribution ($\sigma\!=\!0$), abs.\ diff.\ is 0.955.
Performance improves monotonically as $\sigma$ increases, reaching the best
overall result of 0.912 at $\sigma\!=\!3.6$---the lowest abs.\ diff.\ across
all experiments.
The curve is flat beyond $\sigma\!\approx\!2.0$, so any value in the range
$2$--$4$ yields comparable performance.
Gaussian smoothing approximates the overall distribution shape by spreading
weight toward neighbouring label values, naturally boosting rare labels.

\subsection{Experiment 4: Minimum Samples per Label}

Enforcing min-10 yields abs.\ diff.\,=\,0.993, slightly better than real-only
but substantially worse than distribution-matching approaches.
Min-100 yields 1.109, worse than real-only training.
The min-100 failure is instructive: aggressive up-sampling of rare high-count
labels shifts the distribution toward uniformity in the tail, reintroducing
the same problem that makes a uniform distribution harmful.
Min-10 avoids this by providing basic coverage without distorting the overall
shape---consistent with Experiment~2, where slightly over-representing rare
labels helped---though the gap with Gaussian smoothing (0.912) suggests
smoothing confers broader benefits beyond simple up-sampling.

\subsection{Summary}

\begin{figure}[!t]
  \centering
  \includegraphics[width=\columnwidth]{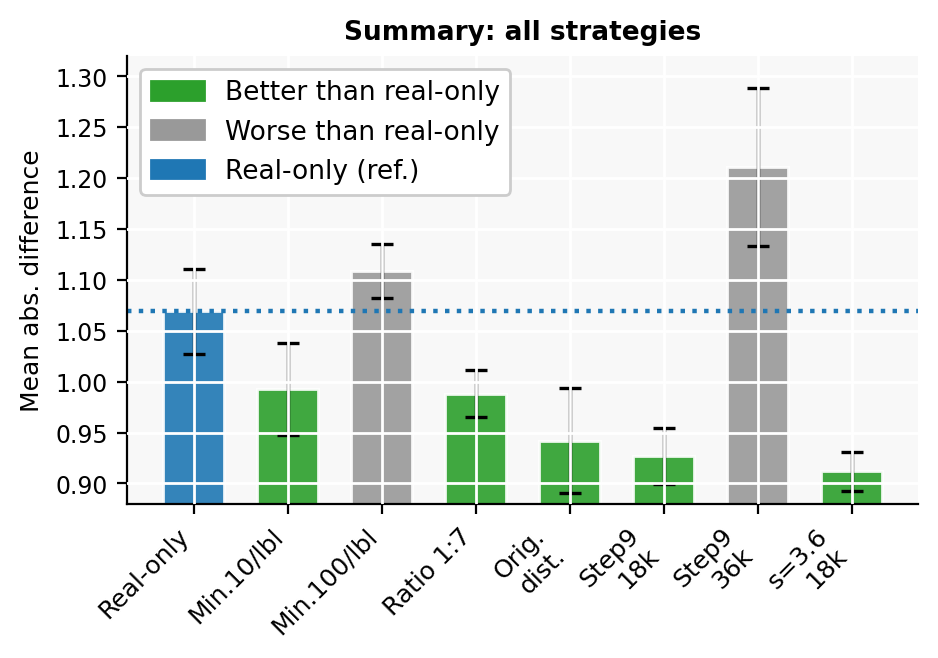}
  \caption{Summary of all strategies.
           Green bars outperform the real-only baseline (dotted line);
           grey bars do not.
           Error bars: $\pm$1\,std, five runs.}
  \label{fig:summary}
\end{figure}

Figure~\ref{fig:summary} compares all approaches.
The best strategy overall is Gaussian smoothing of the real label distribution
($\sigma\!=\!3.6$, abs.\ diff.\,=\,0.912), followed by distribution interpolation
to Step~9 (0.927), the real distribution (0.942), and ratio tuning at
1:7 (0.988).
Note that all distribution experiments used a fixed ratio that was not
independently optimised.
The poor 36k result (Step~9: 1.211) illustrates that even a well-matched
distribution does not compensate for an excess of synthetic data---the
synthetic-to-real ratio remains an important factor alongside distribution
quality.

%%%%%%%%%%%%%%%%%%%%%%%%%%%%%%%%%%%%%%%%%%%%%%%%%%%%%%%%%%%%%%%%%%%%%
\section{Conclusion}
%%%%%%%%%%%%%%%%%%%%%%%%%%%%%%%%%%%%%%%%%%%%%%%%%%%%%%%%%%%%%%%%%%%%%

We systematically evaluated synthetic data design choices for canola branch
counting, covering ratio, label distribution interpolation, Gaussian smoothing,
and minimum-per-label constraints, using a well-calibrated L-system plant
model~\cite{cieslak2021lsystem}.

Three practical recommendations emerge.
First, the synthetic-to-real ratio is broadly forgiving: any ratio in 1:5--1:22
outperforms real-only training, and precise tuning is unnecessary.
Second, label distribution has the largest impact.
Avoiding a uniform distribution and interpolating 90\% toward the real
distribution gives strong results (abs.\ diff.\ 0.927).
Applying Gaussian smoothing ($\sigma\!\approx\!2$--$4$) to the real label
distribution yields the best overall performance (abs.\ diff.\ 0.912,
a 14.7\% improvement over real-only training), showing that simple adjustments
to the observed distribution can improve performance substantially.
Third, minimum-per-label floors offer a simpler alternative to distribution
matching.
Setting the floor too high over-corrects and hurts performance.
A modest minimum of 10 images per label is a safe choice when distribution
resampling is not viable.

Despite requiring labelled data, distribution matching helps close the 
synthetic-to-real gap. The canola branch counting experiments show that 
combined datasets outperform real-only baseline.
Consequently, augmenting real images with synthetic data makes distribution
matching an essential step.

Future work will examine whether these findings generalise to other crops.
Notably, model construction and calibration require substantial effort, and it is
currently unclear which components are most critical for bridging the
synthetic-to-real domain gap.
A systematic investigation of how different aspects of the L-system model
affect machine learning performance would guide future efforts in synthetic
data generation for plant phenotyping.

%%%%%%%%%%%%%%%%%%%%%%%%%%%%%%%%%%%%%%%%%%%%%%%%%%%%%%%%%%%%%%%%%%%%%
\section*{Acknowledgments}
%%%%%%%%%%%%%%%%%%%%%%%%%%%%%%%%%%%%%%%%%%%%%%%%%%%%%%%%%%%%%%%%%%%%%

This work was supported by the Plant Phenotyping and Imaging Research
Centre/Canada First Research Excellence Fund and by the Natural
Sciences and Engineering Research Council of Canada (Discovery Grants
RGPIN-2021-02795 to AR and 2019-06279 to Przemyslaw Prusinkiewicz to support
AD and MC).
We thank Przemyslaw Prusinkiewicz for valuable discussions and input, and the
Debian CI project for the use of hardware provided by AMD.

%%%%%%%%%%%%%%%%%%%%%%%%%%%%%%%%%%%%%%%%%%%%%%%%%%%%%%%%%%%%%%%%%%%%%
\renewcommand\refname{References}
\makeatletter
\let\bib@orig\thebibliography
\def\thebibliography{\bib@orig}
\makeatother
\begin{footnotesize}
\bibliographystyle{unsrt.bst}
\textnormal{\bibliography{references.bib}}
\end{footnotesize}

\end{document}